\documentclass[11pt]{article}

\usepackage[preprint]{acl}

\usepackage{times}
\usepackage{latexsym}
\usepackage[T1]{fontenc}
\usepackage[utf8]{inputenc}
\usepackage{microtype}
\usepackage{inconsolata}
\usepackage{graphicx}
\usepackage{booktabs}
\usepackage{amsmath}
\usepackage{array}
\usepackage{multirow}
\usepackage{listings}
\usepackage{xcolor}
\usepackage{tikz}
\usepackage{placeins}
\usetikzlibrary{positioning, arrows.meta, calc}
\usepackage{color,soul}
\usepackage{cleveref}
\usepackage{float}

\title{Tool-Making and Self-Evolving LLM Agents in Low-Latency Systems}

\author{Kalle Kujanp\"a\"a ~~~ Ning Liu ~~~ Shahnawaz Alam ~~~ Yeshwanth Reddy Sura\\
\textbf{Tianyu Yang ~~~ Kristina Klinkner ~~~ Shervin Malmasi}
\\\\
  Amazon, Fulfillment Technologies \& Robotics \\
  \texttt{\{kujanpaa,malmasi\}@amazon.com} \\}

\begin{document}
\maketitle

\begin{abstract}
Production LLM agents often waste latency and reliability by regenerating code for the same procedural steps on every request. We replace this inference-time coding loop with an agentic tool-making pipeline that compiles repeated SOP steps into validated, versioned tools before deployment. The tool-maker grounds synthesis in the live environment as it collects execution traces, observes backend schemas and values, generates candidate tools, and repairs them against labeled cases. At runtime, the production agent calls these tools directly and falls back to code generation only when needed. We deploy the approach in a Fulfillment Center alarm-triage system, where an agent diagnoses alarms against a 44-node SOP over heterogeneous metric backends. In production, tool calls reduce p50 latency by 42\%. On 1{,}500 historical alarms, they reduce end-to-end error rate by up to 53\% by suppressing run-to-run variance in repeated steps. Because tools return compact structured verdicts, they also enable a simpler direct-call architecture, reducing p50 latency by a further 62\% in a controlled ablation. Versioned tools also improve auditability and expose specification gaps and upstream data drift. Our results show that agents that build and maintain their own tool libraries, a key element of self-evolving agents, can make industrial LLM systems faster, more reliable, and easier to operate.

\end{abstract}

\section{Introduction}

LLM coding agents are becoming a practical interface to complex software and operational systems. Instead of a fixed API or hand-written automation per task, these agents read natural-language instructions, write and execute code, and iterate until they produce an answer or action. This flexibility has enabled deployments in on-call incident triage \citep{chen2023rcacopilot, wang2024rcagent}, workflow-guided operations \citep{ye2025sop}, and general code-based tool use \citep{wang2024executable}.
In industry settings, many procedures are documented as standard operating procedures (SOPs), while the data needed to execute them resides in metric stores, logs, dashboards, and ticketing systems.

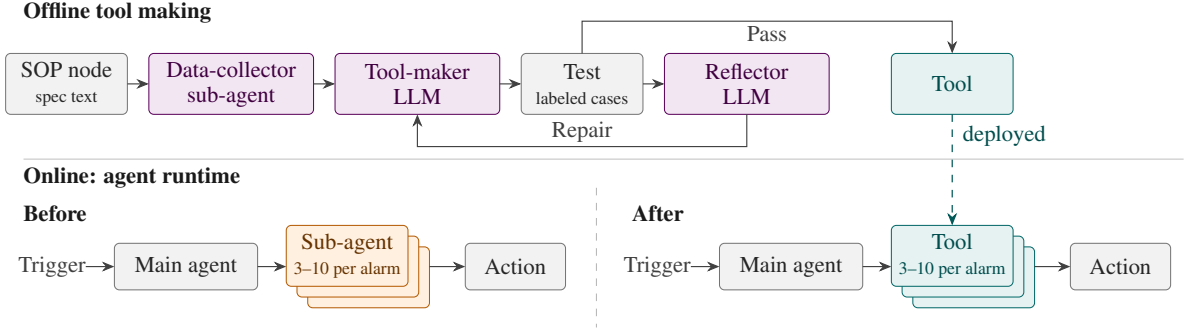
\begin{figure*}[t]
\centering
  \resizebox{0.98\linewidth}{!}{%
\begin{tikzpicture}[
  font=\small,
  >={Stealth[length=5pt,width=4pt]},
  every node/.style={inner sep=0pt, outer sep=0pt},
  graybox/.style={
    rectangle, rounded corners=2pt, line width=0.4pt,
    draw=black!55, fill=black!5, text=black!85,
    align=center, minimum width=1.7cm, minimum height=0.85cm,
    font=\footnotesize
  },
  purplebox/.style={
    rectangle, rounded corners=2pt, line width=0.4pt,
    draw=violet!70!black, fill=violet!10, text=violet!50!black,
    align=center, minimum width=2.3cm, minimum height=0.85cm,
    font=\footnotesize
  },
  tealbox/.style={
    rectangle, rounded corners=2pt, line width=0.4pt,
    draw=teal!80!black, fill=teal!10, text=teal!60!black,
    align=center, minimum width=1.7cm, minimum height=0.85cm,
    font=\footnotesize
  },
  amberbox/.style={
    rectangle, rounded corners=2pt, line width=0.4pt,
    draw=orange!70!black, fill=orange!12, text=orange!40!black,
    align=center, minimum width=1.7cm, minimum height=0.85cm,
    font=\footnotesize
  },
  outputbox/.style={
    rectangle, rounded corners=2pt, line width=0.4pt,
    draw=teal!80!black, fill=teal!10, text=teal!60!black,
    align=center, minimum width=1.7cm, minimum height=0.85cm,
    font=\footnotesize
  },
  agent/.style={
    rectangle, rounded corners=2pt, line width=0.4pt,
    draw=black!55, fill=black!5, text=black!85,
    align=center, minimum width=2.0cm, minimum height=0.65cm,
    font=\footnotesize
  },
  action/.style={
    rectangle, rounded corners=2pt, line width=0.4pt,
    draw=black!55, fill=black!5, text=black!85,
    align=center, minimum width=1.4cm, minimum height=0.65cm,
    font=\footnotesize
  },
  arrow/.style={->, line width=0.45pt, draw=black!75},
  deployedarrow/.style={->, line width=0.5pt, dashed, draw=teal!70!black},
  divider/.style={dashed, line width=0.3pt, draw=black!30},
  caption/.style={font=\footnotesize, text=black!75, align=left},
  sectiontitle/.style={font=\small\bfseries, text=black!90},
  columntitle/.style={font=\small\bfseries, text=black!90},
]

\node[sectiontitle, anchor=west] at (0, -2.30) {Online: agent runtime};

\node[columntitle, anchor=west] at (0, -2.8)   {Before};
\node[columntitle, anchor=west] at (8.5, -2.8) {After};

\node[caption]                       (al1) at (0.4, -3.55) {Trigger};
\node[agent, right=4mm of al1]       (mb1)               {Main agent};
\node[amberbox]                      (s_back) at ($(mb1.east) + (1.55, -0.15)$) {};
\node[amberbox]                      (s_mid)  at ($(mb1.east) + (1.4, 0.0)$)  {};
\node[amberbox]                      (s_front) at ($(mb1.east) + (1.25, 0.15)$) {Sub-agent\\ \scriptsize 3--10 per alarm};
\node[action]                        (act1) at ($(mb1.east) + (3.6, 0)$) {Action};

\draw[arrow] (al1.east) -- (mb1.west);
\draw[arrow] (mb1.east) -- ($(s_front.west) + (0, -0.15)$);
\draw[arrow] ($(s_back.east) + (0, 0.15)$) -- (act1.west);

\draw[divider] (8.0, -2.45) -- (8.0, -4.4);

\node[caption]                       (al2) at (8.85, -3.55) {Trigger};
\node[agent, right=4mm of al2]       (mb3)               {Main agent};
\node[tealbox]                       (n_back) at ($(mb3.east) + (1.55, -0.15)$) {};
\node[tealbox]                       (n_mid)  at ($(mb3.east) + (1.4, 0.0)$)  {};
\node[tealbox]                       (n_front) at ($(mb3.east) + (1.25, 0.15)$) {Tool\\ \scriptsize 3--10 per alarm};
\node[action]                        (act2) at ($(mb3.east) + (3.6, 0)$) {Action};

\draw[arrow] (al2.east) -- (mb3.west);
\draw[arrow] (mb3.east) -- ($(n_front.west) + (0, -0.15)$);
\draw[arrow] ($(n_back.east) + (0, 0.15)$) -- (act2.west);

\draw[line width=0.3pt, draw=black!25] (0, -2.05) -- (16.2, -2.05);

\node[sectiontitle, anchor=west] at (0, 0)
  {Offline tool making};

\node[graybox]                       (sop) at (0.6, -1.0) {SOP node\\ \scriptsize spec text};
\node[purplebox, right=3mm of sop]   (dc)  {Data-collector\\ sub-agent};
\node[purplebox, right=3mm of dc]    (gen) {Tool-maker\\ LLM};
\node[graybox,  right=3mm of gen]    (test){Test\\ \scriptsize labeled cases};
\node[purplebox, right=3mm of test]  (ref) {Reflector\\ LLM};
\node[outputbox]  (nf) at (n_front.center |- 0,-1.0) {Tool};

\draw[arrow] (sop) -- (dc);
\draw[arrow] (dc)  -- (gen);
\draw[arrow] (gen) -- (test);
\draw[arrow] (test) -- (ref);
\draw[arrow] (test.north) |- ($(nf.north) + (0, 0.45)$) -| (nf.north);
\node[below, font=\footnotesize, text=black!75, yshift=-2pt] at ($(test.north)!0.5!(nf.north) + (0, 0.45)$) {Pass};
\draw[arrow] (ref.south) |- ($(gen.south) + (0, -0.45)$) -| (gen.south);
\node[above, font=\footnotesize, text=black!75, yshift=2pt] at ($(gen.south)!0.5!(ref.south) + (0, -0.45)$) {Repair};

\draw[deployedarrow] (nf.south) -- (n_front.north);
\node[caption, text=teal!60!black, anchor=west] at ($(nf.south) + (0.15, -0.30)$) {deployed};

\end{tikzpicture}
}
\caption{%
Offline (top), each SOP node is compiled into a tool: a data-collector sub-agent runs against MCP, producing a trace; a tool-maker LLM writes a candidate from it; and a reflector--tool-maker loop repairs the candidate against labeled cases. Online (bottom), the baseline sub-agent writes fresh MCP-querying code per node over 10--20 LLM turns, whereas our agent calls the compiled tools inline, one call per node.%
}
\label{fig:overview}
\end{figure*}

This flexibility, however, creates an efficiency and reliability bottleneck. In the prevailing CodeAct-style paradigm \citep{wang2024executable}, the agent generates and executes fresh code for each request at inference time. When the same workflow is repeated against a stable backend, the agent re-interprets the same instruction, rediscovers the same schema, and regenerates similar code, raising latency, cost, and run-to-run variance. Prior benchmarks show agents still struggle with workflow-guided tasks \citep{nandi2025sop, wang2025sop, riddell2026stalled}, and even SOP-enhanced multi-agent systems leave a substantial accuracy gap on operational diagnostics \citep{pei2025flow}. In our setting, most latency and correctness errors are caused by translating underspecified SOP text into a concrete query against a production metric backend.

This has motivated a shift from inference-time coding agents to self-evolving agents: systems that improve their own action space over time by constructing reusable tools, skills, or programs before they are needed in production. Recent work explores task-conditioned tool synthesis \citep{cai2024large,yuan2024craft,wang2024trove}, closed-loop construction driven by feedback \citep{wang2023voyager,wolflein2025llm}, reusable workflow induction from demonstrations \citep{wang2024agent}, and the translation of operational documents such as company policies and troubleshooting guides into tools \citep{zwerdling2025towards,mao2025agentic}. The key idea is to amortize cost by identifying procedural steps executed repeatedly and compiling them into deterministic tools, which are then invoked cheaply and consistently at runtime.

We study this approach in a production agent at a Fulfillment Center (FC), where SOPs prescribe how alarms on robotic induct, conveyance, and sortation subsystems are diagnosed and acted upon. These SOPs branch on thresholds from heterogeneous metric backends and select interventions from low-cost flow reallocations to high-cost staffing changes, ticket escalations, and equipment standbys. Latency and correctness are tightly coupled: failures cascade across subsystems within minutes, while an expensive action taken against the wrong cause is difficult to reverse. Manual execution by floor operators is slow under concurrent alarms and error-prone given the number of metrics, schemas, and conditional branches.

We report the design and production deployment of an agentic tool-making pipeline for this setting, triaging outbound dock alarms against a written SOP. The system has two modes (Figure~\ref{fig:overview}). At build time, a tool-maker constructs a library of per-SOP-step tools against the production environment, observing schemas and execution traces and validating candidates against labeled cases rather than relying on SOP text alone. At inference time, the agent invokes these tools per alarm, falling back to code generation only when a tool is unavailable or fails. This preserves the flexibility of coding agents while moving repeated, latency-sensitive work into a reusable and auditable tool library.

We make two main contributions: an agentic tool-making pipeline that compiles repeated SOP steps into validated tools, and its integration into a deployed alarm-triage agent, where it cuts p50 latency by 42\%, enables an architectural redesign that cuts it a further 62\% in a controlled ablation, and reduces the end-to-end error rate by up to 53\%. We also report three findings. The data-collection trace and test-repair loop are both necessary. Residual errors then concentrate on underspecified SOP steps, where targeted fixes raise pass@1 from 94.5\% to 99.9\%. Versioned tools with stable inputs and outputs reveal pre-existing upstream inconsistencies that run-to-run variation otherwise hides. Our results suggest that agents that compile their own experience into reusable tools are a practical path for bringing coding-agent flexibility to latency-sensitive industrial workflows.

\section{Production Setting and Task}

An outbound dock is the final operation in a fulfillment center, where packages are sorted and loaded before they leave for downstream logistics. The dock is tightly coupled: a degraded subsystem backs up the operations feeding it and cascades across the dock within minutes, amplifying a localized fault into an outsized delay that threatens delivery commitments. The agent triages alarms such as robotic stations idling due to upstream conveyance blockages, or producing elevated error rates from camera, sensor, label, or packaging defects.

The agent follows an SOP encoded as a decision tree of 44 decision nodes and 19 action nodes. Decision nodes are of three kinds. Observations test whether a metric crosses an alert threshold and serve as alarm entry points, root causes identify the underlying failure, and constraints gate the most expensive actions. Actions range from cheap (process alerts, flow reallocation) to costly (equipment shutdown, staffing changes). Each node specifies which data to query and how to process it.

The agent evaluates observations to localize the failure, reaches candidate root causes, and executes the action. Before shutting down a robot with a faulty suction cup, for example, constraint nodes verify that the shutdown is either safe (remaining capacity covers the item flow) or unavoidable (the error rate makes operation worse).

\section{Method}

\label{sec:baseline}

We build on the hierarchical multi-agent architecture of Eluna \citep{eluna2026} where a main agent traverses the decision tree and delegates each evaluated node to a CodeAct-style sub-agent \citep{wang2024executable} via a free-text question. The sub-agent generates and executes code that queries metrics through MCP (Model Context Protocol), iterating until it returns a free-text answer, which the main agent parses into a verdict (true/false/no data) to inform subsequent traversal. Delegation keeps the multi-turn data-processing context within the sub-agent, and sub-agent calls within a decision-tree layer run in parallel. Per alarm, the agent evaluates 3--10 nodes, and the sub-agent loop can exceed 100 LLM turns in the most complex cases.

The dominant per-alarm cost in this architecture is the sub-agent's loop, as most of its turns translate the same SOP text into the same MCP queries against an unchanging interface. We replace this loop with per-node tools, each a build-time compilation of the loop for one node into a single deterministic call that returns the verdict directly at inference. Tools are model-agnostic and regenerable independently of agent fine-tuning.

\subsection{Tool-Making Pipeline}
\label{sec:gen}

SOPs are authored by humans, and the SOP text omits execution details that a human reader would resolve from domain knowledge or by inspecting the data, such as field names, datatypes, null handling, and boundary-case conventions. These omissions cause persistent failures when a tool-maker is given only the SOP text. The pipeline closes the gap with two grounding mechanisms (Figure~\ref{fig:overview}, top): a \emph{data-collection trace} of real execution against the live schema, and a \emph{test--repair loop} that corrects each candidate against labeled cases. Each tool is a function with a uniform signature mapping a (warehouse, timestamp, context) input to the node's verdict (true, false, or no data) and supporting detail. The context carries site-specific parameters and verdicts from parent nodes. Each node has a labeled training set, consisting of (input, verdict) pairs from an offline labeler whose logic was specified by subject-matter experts (Appendix~\ref{sec:appLabeler} discusses the labeling).

The pipeline has three model components, shown left-to-right in Figure~\ref{fig:overview}. The \emph{data-collector sub-agent} is the baseline CodeAct sub-agent serving as the inference-time fallback, with access to the production MCP. For each node, it is run on three sampled training cases balanced across verdict labels, writing and executing query code and iterating to a final verdict that is graded against the case label. The resulting trace, passed to the tool-maker, contains the query code, the MCP responses and observed schema (field names, datatypes, value ranges), and the graded verdict, supplying the execution details the SOP omits.

The \emph{tool-maker LLM} produces a candidate tool conditioned on the SOP node text, the node's position in the decision tree (its parent and child nodes), and the data-collector's traces. The candidate is then tested against the full labeled training set, and disagreements are passed to the test--repair loop. The \emph{reflector LLM} writes a short diagnosis of the failures, after which the tool-maker does a rewrite conditioned on the diagnosis and the disagreements. The reflector receives the failing (input, expected, predicted) triples and the previous candidate, and the tool-maker receives the same and the reflector's diagnosis. The loop runs for a budget of three rounds and terminates early once a candidate passes the full training set. The candidate with the highest training-set pass rate is deployed. Appendix~\ref{sec:appD} shows an example from a generated tool.

At inference, tools are used in one of two configurations: the main agent delegates a node to a sub-agent that calls the node's tool, or the main agent calls the tool directly (Figure~\ref{fig:overview}, bottom). Each tool returns the verdict, the observed value, the threshold, and a textual explanation. "No data" is reserved for missing metrics, and exceptions trigger fallback to the CodeAct sub-agent of the baseline. Only the affected node reverts to baseline latency and accuracy. The remaining nodes continue using their tools.

\subsection{Agent Fine-Tuning}
\label{sec:ft}

Production latency constraints cap the deployed alarm-triage agent at a small model size, and small off-the-shelf models fail to follow the SOP reliably. We therefore distill a larger teacher into the small student model: the teacher generates trajectories on the alarm-triage task, rejection sampling retains only those whose extracted observations, root causes, and actions all match ground truth \citep{yuan2023scaling}, and the student is fine-tuned on them with LoRA \citep{hu2022lora}. To keep the student robust when tools are missing or fail, the teacher generates trajectories under varied tool availability: in 75\% of trajectories, the generated tools are available but 25\% of tools are randomly disabled per trajectory, and in the remaining 25\% all node tools are removed so the teacher falls back to the sub-agent for every node.

\section{Tool-Making Experiments}
\label{sec:offline}

\begin{figure}[t]
  \centering
  \includegraphics[width=\columnwidth]{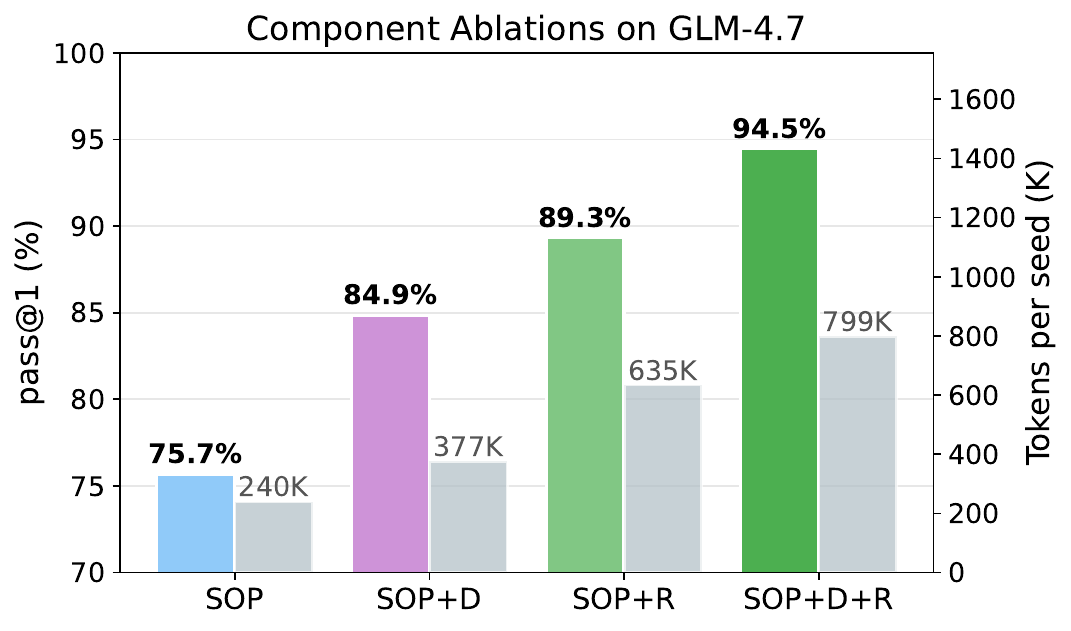}
  \caption{Per-node pass@1 (left) and generation cost in tokens (right) for each ablation configuration on GLM-4.7. +D adds the data-collection trace (Section~\ref{sec:gen}); +R adds the test-repair loop. Only the deployed configuration (SOP+D+R) exceeds 90\% pass@1.}
  \label{fig:cost}
\end{figure}

\begin{table}[t]
  \centering
  \small
  \begin{tabular}{lcccc}
    \toprule
    Tool Maker & SOP & +D & +R & +D+R \\
    \midrule
    GLM-4.7        & 75.7 & 84.9 & 89.3 & \textbf{94.5} \\
    GLM-5          & 76.8 & 83.4 & 88.0 & 93.6 \\
    Qwen3 235B     & 74.2 & 71.6 & 79.9 & 81.1 \\
    GLM-4.7 Flash  & 56.4 & 60.4 & 58.7 & 65.8 \\
    \bottomrule
  \end{tabular}
  \caption{Per-node pass@1 (\%) by tool-maker and configuration. +D adds the data-collection trace; +R adds the test-repair loop. The deployed SOP+D+R column is in bold, with GLM-4.7 deployed. Confidence intervals and significance tests are in Table~\ref{tab:paradigms-ci}.}
  \label{tab:paradigms}
\end{table}

\subsection{Experimental Setup}

\paragraph{Data.}
The pipeline is evaluated on 44 decision nodes. Each node has a labeled training set used by the pipeline and a separate held-out evaluation set. Per-node case counts range from 100 to 200, totaling 8{,}011 training cases and 8{,}022 evaluation cases. Data labels are evenly distributed.

\paragraph{Models.}
We compare the following four models as the tool-making agent (Section~\ref{sec:gen}): GLM-4.7 \citep{zeng2025glm}, GLM-5 \citep{zeng2026glm}, Qwen3 235B \citep{yang2025qwen3}, and GLM-4.7 Flash \citep{zeng2025glm}. We use Claude Opus 4.6 for additional ablations \citep{anthropic2026claudeopus46} to prevent model capacity from confounding the comparison. When reflection is used, the reflector LLM is the same as the tool-maker LLM. The deployed alarm-triage agent uses a separately fine-tuned model (Section~\ref{sec:ft}).

\paragraph{Evaluation.}
We evaluate the build-time pipeline by per-node pass@1: the probability that one full execution of the pipeline produces a tool that passes the held-out evaluation set for that node, where a tool passes only if it returns the correct verdict on every case in the set. Pass@1 is averaged across the 44 decision nodes. Confidence intervals and significance tests are reported in Appendix~\ref{sec:appA}.

\subsection{Main Results}
\label{sec:ablations}

Figure~\ref{fig:cost} shows the per-node pass@1 and generation cost of each configuration on GLM-4.7, and Table~\ref{tab:paradigms} reports pass@1 for all four tool-makers. The full configuration adds the data-collection trace (+D) and the test-repair loop (+R) to the SOP text, reaching 94.5\% pass@1 on GLM-4.7 and 93.6\% on GLM-5. Both components are necessary: removing the repair loop reduces pass@1 by 9.6pp and removing the trace by 5.2pp, and only the full SOP+D+R configuration exceeds 90\%. The two strong tool-makers, GLM-4.7 and GLM-5, behave alike, with the trace and the repair loop each adding a significant gain, whereas the weaker Qwen3 235B and GLM-4.7 Flash gain less and do not improve monotonically. We deploy GLM-4.7, which attains the highest pass@1 of the four. The appendices report further results (Appendix~\ref{sec:appG}). Repair compute cannot substitute for the trace, whose components each contribute a gain, and reflection-based repair outperforms repairing from raw failures only when the trace is present. The pipeline also approaches the full-data ceiling from a handful of labeled cases per node, with a label-free variant within a few points (Appendix~\ref{sec:appC}).

Across the 30 GLM-4.7 pipeline runs, each generated library contains 39--43 tools that pass their full held-out sets (median and mode: 42 of 44), and 38 of the 44 nodes succeed in all 30 runs. Failures are concentrated in three nodes, which succeed in 19, 3, and 0 of the 30 runs, respectively. When they fail, each misses only one to three fixed edge cases out of roughly 200. Thus, although these tools fail the strict pass@1 criterion, they remain correct on the overwhelming majority of their cases, which allows us to deploy them. The deployed library was produced by a single pipeline run under this configuration and contains 42 of 44 tools that pass all held-out cases.

\subsection{Specification Quality as the Bottleneck}
\label{sec:spec}

At this 94.5\% ceiling, four of the 44 nodes account for almost all remaining errors. Three have edge cases (negative denominators, null MCP values) that neither the SOP nor the training labels cover, so no failing case reaches the repair loop. The fourth has an ambiguous averaging window that the repair loop overfits away from the SOP (Appendix~\ref{sec:appD}). Two undeployed Opus runs confirm the cause: augmenting the training data raises pass@1 to 96.6\%, and clarifying the ambiguous SOP raises it to 99.9\% (Table~\ref{tab:ceiling}). The headline 94.5\% pass@1 in Section~\ref{sec:ablations} is unaffected.

\begin{table}[t]
  \centering
  \small
  \begin{tabular}{lc}
    \toprule
    Configuration & pass@1 \\
    \midrule
    Original SOP, original data & 94.5 \\
    Original SOP, augmented data & 96.6 \\
    Clarified SOP, augmented data & \textbf{99.9} \\
    \bottomrule
  \end{tabular}
  \caption{Effect of targeted specification and data improvements on the four residual nodes (Opus). Augmentation adds labeled cases for the edge-case nodes.}
  \label{tab:ceiling}
\end{table}

\section{Agent Experiments and Deployment}
\label{sec:online}

The deployed alarm-triage agent uses small language models to meet production latency targets and calls the GLM-4.7-generated tool library from Section~\ref{sec:offline}. We report runs with two such models (Qwen3 32B \citep{yang2025qwen3} and GLM-4.5-Air \citep{zeng2025glm}), while the tool-making pipeline (\Cref{sec:gen}) runs offline with stronger models, paying their reasoning cost once per tool and amortizing it across every production call.
We evaluate end-to-end correctness on 1{,}500 historical alarm scenarios with ground-truth observations, root causes, and actions, and latency from production. The evaluation alarms are separated from the fine-tuning trajectories (Section~\ref{sec:ft}) by a temporal split and are disjoint from the cases used for tool generation. Non-deployed configurations are evaluated by offline ablation. Before a candidate agent is promoted to production, it runs in shadow: it executes on live alarms in parallel with the production agent without affecting outcomes, and discrepancies between the two are reviewed by humans. Since deployment, our tools have been invoked across over 5{,}000 production alarms.

\subsection{Latency}
\label{sec:latency}

\begin{table}[t]
  \centering
  \small
  \begin{tabular}{llrr}
    \toprule
    Model & Configuration & p50 & p99 \\
    \midrule
    Qwen3 32B & Sub-agent writes code & 100 & 100 \\
    Qwen3 32B & Sub-agent calls tools & 58 & 59 \\
    GLM-4.5-Air & Main-agent direct calls & 26 & 29 \\
    \bottomrule
  \end{tabular}
  \caption{Production p50/p99 latency, relative to the no-tools sub-agent baseline (set to 100, lower is better). Rows 1 and 2 differ only in whether the agent uses tools. Row 3 changes both the architecture (direct calls) and the model (GLM-4.5-Air). A controlled ablation isolates the architecture-only effect, a 62\% p50 reduction (Section~\ref{sec:latency}).}
  \label{tab:latency}
\end{table}

Table~\ref{tab:latency} reports production p50 and p99 latency across three configurations. Replacing the sub-agent's CodeAct loop with tool calls cuts per-alarm production p50 by 42\% on Qwen3 32B under the original two-tier sub-agent architecture (rows 1--2). The same comparison reduces per-alarm output tokens by 58\% (14{,}998 to 6{,}225) and sub-agent turns by 45\% (33.7 to 18.7). A per-component breakdown (Appendix~\ref{sec:appE}) shows the reduction is dominated by an 80\% fall in sub-agent code-generation tokens, with main-agent tokens also falling.

Because tool outputs are small (a few tokens), reading them inline costs little, so the main agent can call tools without a sub-agent, removing the original rationale for the sub-agent layer (\Cref{sec:baseline}). This direct-call configuration's p50 is a further 55\% below row 2, but it also runs GLM-4.5-Air, confounding the two effects. To isolate the architecture, we run an offline ablation that holds the model and inference framework fixed (GLM-4.5-Air, three seeds, 1{,}500 alarms): direct calls alone reduce p50 by 62\% and p99 by 30\% relative to the sub-agent architecture.

\subsection{End-to-End Agent Evaluation}

\begin{table}[t]
  \centering
  \small
  \begin{tabular}{lccc}
    \toprule
     & No tools & Sub-agent & Main-agent \\
    \midrule
    Qwen3 32B & 97.2 & 98.2$^{**}$ & --- \\
    GLM-4.5-Air & 98.3 & 99.2$^{***}$ & 99.3$^{***}$ \\
    \bottomrule
  \end{tabular}
  \caption{End-to-end agent accuracy (\%, $N{=}1,500$). "No tools": the sub-agent writes MCP code, fine-tuned without tools. "Sub-agent" and "Main-agent" both use generated tools, the latter called directly by the main agent. Significance vs.\ no-tools baseline: $^{**}p<0.01$, $^{***}p<0.001$ (Tables~\ref{tab:a4qwen}--\ref{tab:a4glmdirect}).}
  \label{tab:arch}
\end{table}

Table~\ref{tab:arch} reports end-to-end agent accuracy on the 1{,}500 historical alarms across the same three configurations. Tools reduce the error rate from 2.8\% to 1.8\% on Qwen3 32B and from 1.7\% to 0.8\% on GLM-4.5-Air, relative reductions of 36\% and 53\%. Manual inspection of trajectories on which the sub-agent and tool configurations disagree reveals two recurring sub-agent failure modes. First, it deviates from the SOP, using a similar but different metric or aggregating where the SOP specifies a per-entity check. Second, it reaches the correct conclusion but reports it incorrectly to the main agent. Deterministic tools eliminate both.

The agent is fine-tuned on a mixture of trajectories with and without tools (Section~\ref{sec:ft}) so that it can fall back to manual reasoning when a tool is unavailable. When the Qwen3 32B agent fine-tuned with tools is evaluated with tools disabled, its action accuracy (97.4\%) and overall accuracy (96.9\%) are statistically indistinguishable from the 97.4\% and 97.2\% obtained without tools, so mixed-trajectory fine-tuning preserves the fallback.

\section{Discussion and Production Insights}
\label{sec:lessons}
\paragraph{Tools transfer across agents and models with a one-time generation cost.} The library generated by GLM-4.7 serves both the Qwen3 32B and GLM-4.5-Air agents without regenerating any tool. Generating the full 44-node library is a one-time cost of $800$K tokens (Figure~\ref{fig:cost}).

\paragraph{Parameterized tools cover a family of deployments.} A generated tool generalizes across deployments when the quantities that vary between them are tool inputs. In our case these are numeric thresholds, which differ across physical sites with alarm preferences and baseline robot error rates. Each tool resolves them from a runtime parameter store, so one library can serve multiple sites.

\paragraph{Logging tool outputs makes upstream data and environment inconsistencies visible.} Because a tool returns the same value on the same input, its logs reveal environment issues that run-to-run variation in generated code hides.
Every invocation logs the tool version, inputs, and outputs, and a monitoring agent reviews these logs in batch, surfacing drift that no single trajectory would reveal. A flagged tool is corrected by rerunning the generate-test-repair loop and promoting a new version once it passes the held-out set and manual review. Because the agent is decoupled from any specific tool version, deploying or reverting a fix is a configuration change requiring no retraining, and no rollback has been needed since deployment.

This mechanism has revealed three pre-existing inconsistencies in the upstream environment. First, one field differed in type between the offline and production MCPs. Generated code silently adapted, whereas the tool raised a reproducible exception that localized the issue. Second, a node's MCP reported cumulative events while the SOP referred to currently active ones: the tool passed offline evaluation, and the monitoring agent flagged the resulting high positive rate in production. Third, an endpoint silently changed its return format from percentages (75) to decimals (0.75), which the monitoring agent caught from the shifted value distribution.

\section{Related Work}
\label{sec:related}

\paragraph{Tool use and tool-making.} Early LLM tool use focused on calling fixed APIs \citep{schick2023toolformer, patil2024gorilla}. A parallel line gave agents an action space of executable code \citep{gao2023pal, yao2022react, wang2024executable}. A more recent line of work has the agent construct its own tools. These include methods that synthesize tools from task descriptions or datasets \citep{cai2024large,qian2023creator,yuan2024craft,wang2024trove} and agents that iteratively generate and refine tools through repeated execution and testing \citep{wang2023voyager,zheng2025skillweaver,wolflein2025llm}. Operational documents can also be transformed into tools \citep{zwerdling2025towards, mao2025agentic, ye2025sop}. Recent research confirms that prompting-based agents fail to follow SOPs reliably when evaluated against the original SOP code as a verifier \citep{li2025sopbench, nandi2025sop}, with recurring failure modes including route blindness and calculation errors \citep{wang2025sop}. Recent industrial deployments include frameworks for alert triage and root cause analysis in production systems \citep{liu2026bian, bharadwaj2026agentic}. The closest peer work for SOP-to-code is \citet{pei2025flow}, but it is only evaluated on a benchmark. We contribute to this line of work by compiling SOPs into tools validated against labeled production cases.

\paragraph{Iterative refinement and data grounding.} Our work is closely related to the execution-feedback refinement family \citep{chen2024teaching, zhang2023self, zhong2024debug, ridnik2024code}. \citet{olausson2024self} conclude that the success of self-repair depends on the feedback. Our finding that the pipeline plateaus without environmental grounding is consistent with this. The text-to-SQL literature also finds that schema alone hits a ceiling and grounding in database values is required \citep{li2023can, wang2025mac, cao2026apex}. Our findings on the importance of specification quality for tool generation echo what has been reported earlier in the code and SQL generation literature \citep{wretblad2024understanding, jin2026pervasive}.

\section{Conclusion}

We show that repeated inference-time code generation can be optimized for production LLM agents. By compiling SOP nodes into environment-grounded, validated, and versioned tools, our system moves recurring procedural work out of the latency-critical path while preserving fallback flexibility for failures. In a deployed fulfillment-center alarm-triage agent, this reduces latency and improves end-to-end accuracy.

The broader contribution is an operating pattern, a step toward self-evolving industrial agents: the agent maintains the implementation of its action space, building, testing, and repairing tools offline rather than improvising code on every request. The action space itself, defined by the SOP, and tool promotion remain human decisions. Once grounded tools are in place, the limiting factor becomes the quality of the underlying SOP specification instead of the agent's ability to rediscover schemas or rewrite boilerplate. This makes agent behavior faster, more consistent, and more auditable, suggesting a practical route to reliable low-latency LLM systems in operational workflows.

\section*{Limitations}

Our agentic tool-making loop has been deployed on one application, outbound dock alarm triage. Generalization to other application domains, such as free-form runbooks, is not established. The setting nonetheless shares properties with typical agentic workflows: a written procedure that the agent must follow and that is executed against a stable backend, and latency or auditability constraints that make inference-time code generation less appealing as an alternative.

The improvement loop retains human oversight for production safety. It already automates parts of the cycle, such as drift detection and candidate tool generation, but we cannot yet guarantee that every failure would be caught without human review, so fully automating it remains open. Clarifying an underspecified SOP is inherently a human task, because the SOP encodes the author's intent for how the procedure should behave. Creating a new tool still benefits from labeled cases. Our label-free configuration trails the labeled ceiling by less than 3pp, and a fully label-free pipeline would remove the need for hand-labeled cases when adding new nodes.

\bibliography{custom}

@inproceedings{chen2023rcacopilot,
  title={Automatic Root Cause Analysis via Large Language Models for Cloud Incidents}, 
  author={Yinfang Chen and Huaibing Xie and Minghua Ma and Yu Kang and Xin Gao and Liu Shi and Yunjie Cao and Xuedong Gao and Hao Fan and Ming Wen and Jun Zeng and Supriyo Ghosh and Xuchao Zhang and Chaoyun Zhang and Qingwei Lin and Saravan Rajmohan and Dongmei Zhang and Tianyin Xu},
  booktitle={Proceedings of the Nineteenth European Conference on Computer Systems},
  pages={674--688},
  year={2024}
}

@inproceedings{wang2024rcagent,
  title={Rcagent: Cloud root cause analysis by autonomous agents with tool-augmented large language models},
  author={Wang, Zefan and Liu, Zichuan and Zhang, Yingying and Zhong, Aoxiao and Wang, Jihong and Yin, Fengbin and Fan, Lunting and Wu, Lingfei and Wen, Qingsong},
  booktitle={Proceedings of the 33rd ACM international conference on information and knowledge management},
  pages={4966--4974},
  year={2024}
}

@article{ye2025sop,
  title={Sop-agent: Empower general purpose ai agent with domain-specific sops},
  author={Ye, Anbang and Ma, Qianran and Chen, Jia and Li, Muqi and Li, Tong and Liu, Fujiao and Mai, Siqi and Lu, Meichen and Bao, Haitao and You, Yang},
  journal={arXiv preprint arXiv:2501.09316},
  year={2025}
}

@inproceedings{wang2024executable,
  title={Executable code actions elicit better llm agents},
  author={Wang, Xingyao and Chen, Yangyi and Yuan, Lifan and Zhang, Yizhe and Li, Yunzhu and Peng, Hao and Ji, Heng},
  booktitle={Forty-first International Conference on Machine Learning},
  year={2024}
}

@article{nandi2025sop,
      title={SOP-Bench: Complex Industrial SOPs for Evaluating LLM Agents}, 
      author={Subhrangshu Nandi and Arghya Datta and Rohith Nama and Udita Patel and Nikhil Vichare and Indranil Bhattacharya and Prince Grover and Shivam Asija and Giuseppe Carenini and Wei Zhang and Arushi Gupta and Sreyoshi Bhaduri and Jing Xu and Huzefa Raja and Shayan Ray and Aaron Chan and Esther Xu Fei and Gaoyuan Du and Zuhaib Akhtar and Harshita Asnani and Weian Chan and Ming Xiong and Francesco Carbone and Jeetu Mirchandani},
  journal={arXiv preprint arXiv:2506.08119},
  year={2025}
}

@article{wang2025sop,
  title={SOP-Maze: Evaluating Large Language Models on Complicated Business Standard Operating Procedures},
  author={Wang, Jiaming and Tang, Zhe and Jin, Zehao and Chen, Hefei and Jin, Yilin and Ding, Peng and Li, Xiaoyu and Cao, Xuezhi},
  journal={arXiv preprint arXiv:2510.08942},
  year={2025}
}

@article{riddell2026stalled,
  title={Stalled, Biased, and Confused: Uncovering Reasoning Failures in LLMs for Cloud-Based Root Cause Analysis},
  author={Riddell, Evelien and Riddell, James and Sun, Gengyi and Antkiewicz, Micha{\l} and Czarnecki, Krzysztof},
  journal={arXiv preprint arXiv:2601.22208},
  year={2026}
}

@inproceedings{pei2025flow,
      title={Flow-of-Action: SOP Enhanced LLM-Based Multi-Agent System for Root Cause Analysis}, 
      author={Changhua Pei and Zexin Wang and Fengrui Liu and Zeyan Li and Yang Liu and Xiao He and Rong Kang and Tieying Zhang and Jianjun Chen and Jianhui Li and Gaogang Xie and Dan Pei},
  booktitle={Companion Proceedings of the ACM on Web Conference 2025},
  pages={422--431},
  year={2025}
}

@inproceedings{qian2023creator,
    title = "{CREATOR}: Tool Creation for Disentangling Abstract and Concrete Reasoning of Large Language Models",
    author = "Qian, Cheng  and
      Han, Chi  and
      Fung, Yi  and
      Qin, Yujia  and
      Liu, Zhiyuan  and
      Ji, Heng",
    editor = "Bouamor, Houda  and
      Pino, Juan  and
      Bali, Kalika",
    booktitle = "Findings of the Association for Computational Linguistics: EMNLP 2023",
    month = dec,
    year = "2023",
    address = "Singapore",
    publisher = "Association for Computational Linguistics",
    url = "https://aclanthology.org/2023.findings-emnlp.462/",
    doi = "10.18653/v1/2023.findings-emnlp.462",
    pages = "6922--6939"
}

@inproceedings{cai2024large,
  title={Large language models as tool makers},
  author={Cai, Tianle and Wang, Xuezhi and Ma, Tengyu and Chen, Xinyun and Zhou, Denny},
  booktitle={International Conference on Learning Representations},
  year={2024}
}

@inproceedings{yuan2024craft,
  title={Craft: Customizing llms by creating and retrieving from specialized toolsets},
  author={Yuan, Lifan and Chen, Yangyi and Wang, Xingyao and Fung, Yi and Peng, Hao and Ji, Heng},
  booktitle={International Conference on Learning Representations},
  year={2024}
}

@article{wang2024trove,
  title={Trove: Inducing verifiable and efficient toolboxes for solving programmatic tasks},
  author={Wang, Zhiruo and Fried, Daniel and Neubig, Graham},
  journal={arXiv preprint arXiv:2401.12869},
  year={2024}
}

@article{wang2023voyager,
  title={Voyager: An open-ended embodied agent with large language models},
  author={Wang, Guanzhi and Xie, Yuqi and Jiang, Yunfan and Mandlekar, Ajay and Xiao, Chaowei and Zhu, Yuke and Fan, Linxi and Anandkumar, Anima},
  journal={arXiv preprint arXiv:2305.16291},
  year={2023}
}

@article{wang2024agent,
  title={Agent workflow memory},
  author={Wang, Zora Zhiruo and Mao, Jiayuan and Fried, Daniel and Neubig, Graham},
  journal={arXiv preprint arXiv:2409.07429},
  year={2024}
}

@article{zheng2025skillweaver,
      title={SkillWeaver: Web Agents can Self-Improve by Discovering and Honing Skills}, 
      author={Boyuan Zheng and Michael Y. Fatemi and Xiaolong Jin and Zora Zhiruo Wang and Apurva Gandhi and Yueqi Song and Yu Gu and Jayanth Srinivasa and Gaowen Liu and Graham Neubig and Yu Su},
  journal={arXiv preprint arXiv:2504.07079},
  year={2025}
}

@inproceedings{wolflein2025llm,
  title={Llm agents making agent tools},
  author={W{\"o}lflein, Georg and Ferber, Dyke and Truhn, Daniel and Arandjelovic, Ognjen and Kather, Jakob Nikolas},
  booktitle={Proceedings of the 63rd Annual Meeting of the Association for Computational Linguistics (Volume 1: Long Papers)},
  pages={26092--26130},
  year={2025}
}

@inproceedings{zwerdling2025towards,
    title = "Towards Enforcing Company Policy Adherence in Agentic Workflows",
    author = "Zwerdling, Naama  and
      Boaz, David  and
      Rabinovich, Ella  and
      Uziel, Guy  and
      Amid, David  and
      Anaby Tavor, Ateret",
    editor = "Potdar, Saloni  and
      Rojas-Barahona, Lina  and
      Montella, Sebastien",
    booktitle = "Proceedings of the 2025 Conference on Empirical Methods in Natural Language Processing: Industry Track",
    month = nov,
    year = "2025",
    address = "Suzhou (China)",
    publisher = "Association for Computational Linguistics",
    url = "https://aclanthology.org/2025.emnlp-industry.41/",
    doi = "10.18653/v1/2025.emnlp-industry.41",
    pages = "595--606",
    ISBN = "979-8-89176-333-3"
}

@article{mao2025agentic,
  title={Stepfly: Agentic troubleshooting guide automation for incident diagnosis},
  author={Jiayi Mao and Liqun Li and Yanjie Gao and Zegang Peng and Shilin He and Chaoyun Zhang and Si Qin and Samia Khalid and Qingwei Lin and Saravan Rajmohan and Sitaram Lanka and Dongmei Zhang},
  journal={Proceedings of the ACM on Software Engineering},
  volume={3},
  number={FSE},
  pages={3070--3092},
  year={2026},
  publisher={ACM New York, NY, USA}
}

@article{zeng2025glm,
      title={GLM-4.5: Agentic, Reasoning, and Coding (ARC) Foundation Models}, 
      author={Aohan Zeng and Xin Lv and Qinkai Zheng and Zhenyu Hou and Bin Chen and Chengxing Xie and Cunxiang Wang and Da Yin and Hao Zeng and Jiajie Zhang and Kedong Wang and Lucen Zhong and Mingdao Liu and Rui Lu and Shulin Cao and Xiaohan Zhang and Xuancheng Huang and Yao Wei and Yean Cheng and Yifan An and Yilin Niu and Yuanhao Wen and Yushi Bai and Zhengxiao Du and Zihan Wang and Zilin Zhu and Bohan Zhang and Bosi Wen and Bowen Wu and Bowen Xu and Can Huang and Casey Zhao and Changpeng Cai and Chao Yu and Chen Li and Chendi Ge and Chenghua Huang and Chenhui Zhang and Chenxi Xu and Chenzheng Zhu and Chuang Li and Congfeng Yin and Daoyan Lin and Dayong Yang and Dazhi Jiang and Ding Ai and Erle Zhu and Fei Wang and Gengzheng Pan and Guo Wang and Hailong Sun and Haitao Li and Haiyang Li and Haiyi Hu and Hanyu Zhang and Hao Peng and Hao Tai and Haoke Zhang and Haoran Wang and Haoyu Yang and He Liu and He Zhao and Hongwei Liu and Hongxi Yan and Huan Liu and Huilong Chen and Ji Li and Jiajing Zhao and Jiamin Ren and Jian Jiao and Jiani Zhao and Jianyang Yan and Jiaqi Wang and Jiayi Gui and Jiayue Zhao and Jie Liu and Jijie Li and Jing Li and Jing Lu and Jingsen Wang and Jingwei Yuan and Jingxuan Li and Jingzhao Du and Jinhua Du and Jinxin Liu and Junkai Zhi and Junli Gao and Ke Wang and Lekang Yang and Liang Xu and Lin Fan and Lindong Wu and Lintao Ding and Lu Wang and Man Zhang and Minghao Li and Minghuan Xu and Mingming Zhao and Mingshu Zhai and Pengfan Du and Qian Dong and Shangde Lei and Shangqing Tu and Shangtong Yang and Shaoyou Lu and Shijie Li and Shuang Li and Shuang-Li and Shuxun Yang and Sibo Yi and Tianshu Yu and Wei Tian and Weihan Wang and Wenbo Yu and Weng Lam Tam and Wenjie Liang and Wentao Liu and Xiao Wang and Xiaohan Jia and Xiaotao Gu and Xiaoying Ling and Xin Wang and Xing Fan and Xingru Pan and Xinyuan Zhang and Xinze Zhang and Xiuqing Fu and Xunkai Zhang and Yabo Xu and Yandong Wu and Yida Lu and Yidong Wang and Yilin Zhou and Yiming Pan and Ying Zhang and Yingli Wang and Yingru Li and Yinpei Su and Yipeng Geng and Yitong Zhu and Yongkun Yang and Yuhang Li and Yuhao Wu and Yujiang Li and Yunan Liu and Yunqing Wang and Yuntao Li and Yuxuan Zhang and Zezhen Liu and Zhen Yang and Zhengda Zhou and Zhongpei Qiao and Zhuoer Feng and Zhuorui Liu and Zichen Zhang and Zihan Wang and Zijun Yao and Zikang Wang and Ziqiang Liu and Ziwei Chai and Zixuan Li and Zuodong Zhao and Wenguang Chen and Jidong Zhai and Bin Xu and Minlie Huang and Hongning Wang and Juanzi Li and Yuxiao Dong and Jie Tang},
  journal={arXiv preprint arXiv:2508.06471},
  year={2025}
}

@inproceedings{olausson2024self,
  title={Is self-repair a silver bullet for code generation?},
  author={Olausson, Theo X and Inala, Jeevana Priya and Wang, Chenglong and Gao, Jianfeng and Solar-Lezama, Armando},
  booktitle={International Conference on Learning Representations},
  year={2024}
}

@article{schick2023toolformer,
  title={Toolformer: Language models can teach themselves to use tools},
  author={Schick, Timo and Dwivedi-Yu, Jane and Dess{\`\i}, Roberto and Raileanu, Roberta and Lomeli, Maria and Hambro, Eric and Zettlemoyer, Luke and Cancedda, Nicola and Scialom, Thomas},
  journal={Advances in neural information processing systems},
  volume={36},
  pages={68539--68551},
  year={2023}
}

@article{patil2024gorilla,
  title={Gorilla: Large language model connected with massive apis},
  author={Patil, Shishir G and Zhang, Tianjun and Wang, Xin and Gonzalez, Joseph E},
  journal={Advances in Neural Information Processing Systems},
  volume={37},
  pages={126544--126565},
  year={2024}
}

@inproceedings{gao2023pal,
  title={Pal: Program-aided language models},
  author={Gao, Luyu and Madaan, Aman and Zhou, Shuyan and Alon, Uri and Liu, Pengfei and Yang, Yiming and Callan, Jamie and Neubig, Graham},
  booktitle={International conference on machine learning},
  pages={10764--10799},
  year={2023},
  organization={PMLR}
}

@article{yao2022react,
  title={React: Synergizing reasoning and acting in language models},
  author={Yao, Shunyu and Zhao, Jeffrey and Yu, Dian and Du, Nan and Shafran, Izhak and Narasimhan, Karthik and Cao, Yuan},
  journal={arXiv preprint arXiv:2210.03629},
  year={2022}
}

@article{li2025sopbench,
      title={SOPBench: Evaluating Language Agents at Following Standard Operating Procedures and Constraints}, 
      author={Zekun Li and Shinda Huang and Jiangtian Wang and Nathan Zhang and Antonis Antoniades and Wenyue Hua and Kaijie Zhu and Sirui Zeng and Chi Wang and William Yang Wang and Xifeng Yan},
  journal={arXiv preprint arXiv:2503.08669},
  year={2025}
}

@article{bharadwaj2026agentic,
  title={Agentic Observability: Automated Alert Triage for Adobe E-Commerce},
  author={Bharadwaj, Aprameya and Tu, Kyle},
  journal={arXiv preprint arXiv:2602.02585},
  year={2026}
}

@article{liu2026bian,
      title={Bian Que: An Agentic Framework with Flexible Skill Arrangement for Online System Operations}, 
      author={Bochao Liu and Zhipeng Qian and Yang Zhao and Xinyuan Jiang and Zihan Liang and Yufei Ma and Junpeng Zhuang and Ben Chen and Shuo Yang and Hongen Wan and Yao Wu and Chenyi Lei and Xiao Liang},
  journal={arXiv preprint arXiv:2604.26805},
  year={2026}
}

@inproceedings{chen2024teaching,
  title={Teaching large language models to self-debug},
  author={Chen, Xinyun and Lin, Maxwell and Sch{\"a}rli, Nathanael and Zhou, Denny},
  booktitle={International Conference on Learning Representations},
  year={2024}
}

@inproceedings{zhang2023self,
  title={Self-edit: Fault-aware code editor for code generation},
  author={Zhang, Kechi and Li, Zhuo and Li, Jia and Li, Ge and Jin, Zhi},
  booktitle={Proceedings of the 61st Annual Meeting of the Association for Computational Linguistics (Volume 1: Long Papers)},
  pages={769--787},
  year={2023}
}

@inproceedings{zhong2024debug,
    title = "Debug like a Human: A Large Language Model Debugger via Verifying Runtime Execution Step by Step",
    author = "Zhong, Li  and
      Wang, Zilong  and
      Shang, Jingbo",
    editor = "Ku, Lun-Wei  and
      Martins, Andre  and
      Srikumar, Vivek",
    booktitle = "Findings of the Association for Computational Linguistics: ACL 2024",
    month = aug,
    year = "2024",
    address = "Bangkok, Thailand",
    publisher = "Association for Computational Linguistics",
    url = "https://aclanthology.org/2024.findings-acl.49/",
    doi = "10.18653/v1/2024.findings-acl.49",
    pages = "851--870"
}

@article{ridnik2024code,
  title={Code generation with alphacodium: From prompt engineering to flow engineering},
  author={Ridnik, Tal and Kredo, Dedy and Friedman, Itamar},
  journal={arXiv preprint arXiv:2401.08500},
  year={2024}
}

@article{li2023can,
      title={Can LLM Already Serve as A Database Interface? A BIg Bench for Large-Scale Database Grounded Text-to-SQLs}, 
      author={Jinyang Li and Binyuan Hui and Ge Qu and Jiaxi Yang and Binhua Li and Bowen Li and Bailin Wang and Bowen Qin and Rongyu Cao and Ruiying Geng and Nan Huo and Xuanhe Zhou and Chenhao Ma and Guoliang Li and Kevin C. C. Chang and Fei Huang and Reynold Cheng and Yongbin Li},
  journal={Advances in Neural Information Processing Systems},
  volume={36},
  pages={42330--42357},
  year={2023}
}

@inproceedings{wang2025mac,
      title={MAC-SQL: A Multi-Agent Collaborative Framework for Text-to-SQL}, 
      author={Bing Wang and Changyu Ren and Jian Yang and Xinnian Liang and Jiaqi Bai and LinZheng Chai and Zhao Yan and Qian-Wen Zhang and Di Yin and Xing Sun and Zhoujun Li},
  booktitle={Proceedings of the 31st International Conference on Computational Linguistics},
  pages={540--557},
  year={2025}
}

@inproceedings{cao2026apex,
  title={APEX-SQL: Talking to the data via Agentic Exploration for Text-to-SQL},
  author={Cao, Bowen and Liao, Weibin and Sun, Yushi and Fang, Dong and Li, Haitao and Lam, Wai},
  booktitle={Proceedings of the 32nd ACM SIGKDD Conference on Knowledge Discovery and Data Mining V. 2},
  pages={234--244},
  year={2026}
}

@inproceedings{wretblad2024understanding,
  title={Understanding the effects of noise in text-to-sql: An examination of the bird-bench benchmark},
  author={Wretblad, Niklas and Riseby, Fredrik and Biswas, Rahul and Ahmadi, Amin and Holmstr{\"o}m, Oskar},
  booktitle={Proceedings of the 62nd Annual Meeting of the Association for Computational Linguistics (Volume 2: Short Papers)},
  pages={356--369},
  year={2024}
}

@article{jin2026pervasive,
  title={Pervasive Annotation Errors Break Text-to-SQL Benchmarks and Leaderboards},
  author={Jin, Tengjun and Choi, Yoojin and Zhu, Yuxuan and Kang, Daniel},
  journal={arXiv preprint arXiv:2601.08778},
  year={2026}
}

@article{yuan2023scaling,
  title={Scaling relationship on learning mathematical reasoning with large language models},
  author={Yuan, Zheng and Yuan, Hongyi and Li, Chengpeng and Dong, Guanting and Lu, Keming and Tan, Chuanqi and Zhou, Chang and Zhou, Jingren},
  journal={arXiv preprint arXiv:2308.01825},
  year={2023}
}

@article{zeng2026glm,
      title={GLM-5: from Vibe Coding to Agentic Engineering}, 
      author={Aohan Zeng and Xin Lv and Zhenyu Hou and Zhengxiao Du and Qinkai Zheng and Bin Chen and Da Yin and Chendi Ge and Chenghua Huang and Chengxing Xie and Chenzheng Zhu and Congfeng Yin and Cunxiang Wang and Gengzheng Pan and Hao Zeng and Haoke Zhang and Haoran Wang and Huilong Chen and Jiajie Zhang and Jian Jiao and Jiaqi Guo and Jingsen Wang and Jingzhao Du and Jinzhu Wu and Kedong Wang and Lei Li and Lin Fan and Lucen Zhong and Mingdao Liu and Mingming Zhao and Pengfan Du and Qian Dong and Rui Lu and Shuang-Li and Shulin Cao and Song Liu and Ting Jiang and Xiaodong Chen and Xiaohan Zhang and Xuancheng Huang and Xuezhen Dong and Yabo Xu and Yao Wei and Yifan An and Yilin Niu and Yitong Zhu and Yuanhao Wen and Yukuo Cen and Yushi Bai and Zhongpei Qiao and Zihan Wang and Zikang Wang and Zilin Zhu and Ziqiang Liu and Zixuan Li and Bojie Wang and Bosi Wen and Can Huang and Changpeng Cai and Chao Yu and Chen Li and Chengwei Hu and Chenhui Zhang and Dan Zhang and Daoyan Lin and Dayong Yang and Di Wang and Ding Ai and Erle Zhu and Fangzhou Yi and Feiyu Chen and Guohong Wen and Hailong Sun and Haisha Zhao and Haiyi Hu and Hanchen Zhang and Hanrui Liu and Hanyu Zhang and Hao Peng and Hao Tai and Haobo Zhang and He Liu and Hongwei Wang and Hongxi Yan and Hongyu Ge and Huan Liu and Huanpeng Chu and Jia'ni Zhao and Jiachen Wang and Jiajing Zhao and Jiamin Ren and Jiapeng Wang and Jiaxin Zhang and Jiayi Gui and Jiayue Zhao and Jijie Li and Jing An and Jing Li and Jingwei Yuan and Jinhua Du and Jinxin Liu and Junkai Zhi and Junwen Duan and Kaiyue Zhou and Kangjian Wei and Ke Wang and Keyun Luo and Laiqiang Zhang and Leigang Sha and Liang Xu and Lindong Wu and Lintao Ding and Lu Chen and Minghao Li and Nianyi Lin and Pan Ta and Qiang Zou and Rongjun Song and Ruiqi Yang and Shangqing Tu and Shangtong Yang and Shaoxiang Wu and Shengyan Zhang and Shijie Li and Shuang Li and Shuyi Fan and Wei Qin and Wei Tian and Weining Zhang and Wenbo Yu and Wenjie Liang and Xiang Kuang and Xiangmeng Cheng and Xiangyang Li and Xiaoquan Yan and Xiaowei Hu and Xiaoying Ling and Xing Fan and Xingye Xia and Xinyuan Zhang and Xinze Zhang and Xirui Pan and Xu Zou and Xunkai Zhang and Yadi Liu and Yandong Wu and Yanfu Li and Yidong Wang and Yifan Zhu and Yijun Tan and Yilin Zhou and Yiming Pan and Ying Zhang and Yinpei Su and Yipeng Geng and Yong Yan and Yonglin Tan and Yuean Bi and Yuhan Shen and Yuhao Yang and Yujiang Li and Yunan Liu and Yunqing Wang and Yuntao Li and Yurong Wu and Yutao Zhang and Yuxi Duan and Yuxuan Zhang and Zezhen Liu and Zhengtao Jiang and Zhenhe Yan and Zheyu Zhang and Zhixiang Wei and Zhuo Chen and Zhuoer Feng and Zijun Yao and Ziwei Chai and Ziyuan Wang and Zuzhou Zhang and Bin Xu and Minlie Huang and Hongning Wang and Juanzi Li and Yuxiao Dong and Jie Tang},
  journal={arXiv preprint arXiv:2602.15763},
  year={2026}
}

@misc{anthropic2026claudeopus46,
  author       = {{Anthropic}},
  title        = {Claude Opus 4.6 System Card},
  year         = {2026},
  month        = feb,
  url          = {https://www-cdn.anthropic.com/0dd865075ad3132672ee0ab40b05a53f14cf5288.pdf},
  note         = {Accessed: 2026-06-04}
}

@inproceedings{kwon2023efficient,
  title={Efficient memory management for large language model serving with pagedattention},
  author={Kwon, Woosuk and Li, Zhuohan and Zhuang, Siyuan and Sheng, Ying and Zheng, Lianmin and Yu, Cody Hao and Gonzalez, Joseph and Zhang, Hao and Stoica, Ion},
  booktitle={Proceedings of the 29th symposium on operating systems principles},
  pages={611--626},
  year={2023}
}

@article{yang2025qwen3,
      title={Qwen3 Technical Report}, 
      author={An Yang and Anfeng Li and Baosong Yang and Beichen Zhang and Binyuan Hui and Bo Zheng and Bowen Yu and Chang Gao and Chengen Huang and Chenxu Lv and Chujie Zheng and Dayiheng Liu and Fan Zhou and Fei Huang and Feng Hu and Hao Ge and Haoran Wei and Huan Lin and Jialong Tang and Jian Yang and Jianhong Tu and Jianwei Zhang and Jianxin Yang and Jiaxi Yang and Jing Zhou and Jingren Zhou and Junyang Lin and Kai Dang and Keqin Bao and Kexin Yang and Le Yu and Lianghao Deng and Mei Li and Mingfeng Xue and Mingze Li and Pei Zhang and Peng Wang and Qin Zhu and Rui Men and Ruize Gao and Shixuan Liu and Shuang Luo and Tianhao Li and Tianyi Tang and Wenbiao Yin and Xingzhang Ren and Xinyu Wang and Xinyu Zhang and Xuancheng Ren and Yang Fan and Yang Su and Yichang Zhang and Yinger Zhang and Yu Wan and Yuqiong Liu and Zekun Wang and Zeyu Cui and Zhenru Zhang and Zhipeng Zhou and Zihan Qiu},
  journal={arXiv preprint arXiv:2505.09388},
  year={2025}
}

@inproceedings{hu2022lora,
  title={{LoRA}: Low-Rank Adaptation of Large Language Models},
  author={Hu, Edward J and Shen, Yelong and Wallis, Phillip and Allen-Zhu, Zeyuan and Li, Yuanzhi and Wang, Shean and Wang, Lu and Chen, Weizhu},
  booktitle={International Conference on Learning Representations},
  year={2022}
}

@inproceedings{eluna2026,
  author       = {Ning Liu and P Aditya Sreekar and Kalle Kujanp\"a\"a and Zhaoxuan Zhu and Kaiwen Liu and Chuanneng Sun and Jorge Marchena Menendez and Matthew Bales and Tianyu Yang and Shahnawaz Alam and Rose Yu and Baoyuan Liu and Kristina Klinkner and Shervin Malmasi},
  title        = {Eluna: An Agentic {LLM} System for Automating Warehouse Operations with Reasoning and Task Execution},
  year         = {2026},
    booktitle = "Proceedings of the 2026 Conference on Empirical Methods in Natural Language Processing: Industry Track",
    address = "Budapest (Hungary)",
    publisher = "Association for Computational Linguistics",
}

@misc{zhao2024swiftascalablelightweightinfrastructure,
      title={SWIFT:A Scalable lightWeight Infrastructure for Fine-Tuning},
      author={Yuze Zhao and Jintao Huang and Jinghan Hu and Xingjun Wang and Yunlin Mao and Daoze Zhang and Hong Zhang and Zeyinzi Jiang and Zhikai Wu and Baole Ai and Ang Wang and Wenmeng Zhou and Yingda Chen},
      year={2024},
      eprint={2408.05517},
      archivePrefix={arXiv},
      primaryClass={cs.CL},
      url={https://arxiv.org/abs/2408.05517},
}

\FloatBarrier

\onecolumn

\appendix

\section*{\centering Appendix}

\section{Experimental Setup}
\label{sec:appB}

\paragraph{Seeds.} Per-node pass@1 results in Section~\ref{sec:offline} use 30 independent generation runs per configuration, except the SOP column in Table~\ref{tab:paradigms} and the augmented-data row in Table~\ref{tab:ceiling}, which use 10. End-to-end agent accuracy in Section~\ref{sec:online} uses 3 seeds per configuration on the same 1{,}500 alarm scenarios.

\paragraph{Inference framework.} The deployed agent (Qwen3 32B and GLM-4.5-Air) runs on vLLM \citep{kwon2023efficient} for both offline benchmarks and production. The open-weight tool-makers (GLM-4.7, Qwen3 235B, GLM-4.7 Flash) likewise run on vLLM. GLM-5 and the Claude ablations are served via Amazon Bedrock.

\paragraph{Hardware.} Inference, both offline benchmarks and production, runs on a single 8$\times$H100 instance. Fine-tuning runs on 64 H100s for Qwen3 32B and 128 H100s for GLM-4.5-Air.

\paragraph{Latency measurement.} For competitive reasons we cannot disclose absolute production latencies. Hence, Table~\ref{tab:latency} reports figures normalized to the no-tools sub-agent baseline.

\paragraph{Fine-tuning.} We follow the training procedure of \citet{eluna2026}. We distill GLM-4.7 \citep{zeng2025glm} into the two deployed students, Qwen3 32B and GLM-4.5-Air. A teacher trajectory is retained only when its extracted observations, root causes, and actions all match the ground-truth labels. This filter accepts approximately 92\% of trajectories (93\% on the with-tools slice with dropout, 4{,}171/4{,}500, and 91\% on the no-tools slice, 1{,}358/1{,}500). The same filter applied to the no-tools baseline accepts 91\% of trajectories (6{,}786/7{,}500), so tool availability also raises the teacher's own accuracy. Students are fine-tuned with LoRA \citep{hu2022lora} using MS-SWIFT \citep{zhao2024swiftascalablelightweightinfrastructure}, with rank and $\alpha$ of 256 and learning rate $2{\times}10^{-4}$ for Qwen3 32B and 32 and $1{\times}10^{-4}$ for GLM-4.5-Air. Both train at batch size 64 with a 5\% linear warmup and cosine decay to zero, on 64 and 128 H100 GPUs respectively. The with-tools fine-tunes use one epoch, while the no-tools baseline fine-tunes use two epochs \citep{eluna2026}. After the training, we merge the LoRA adapters so inference carries no added overhead.

\section{Ground-Truth Labeling}
\label{sec:appLabeler}

The pipeline consumes labeled (input, verdict) cases per node (Section~\ref{sec:gen}). Human annotations could provide these labels directly, and only a few labels are needed: pass@1 reaches 91.8\% from five labeled cases per node, and the label-free variant reaches 91.6\% with none (Appendices~\ref{sec:appG} and~\ref{sec:appC}). However, we lacked the annotator capacity to label all the training and evaluation cases used in our experiments, so we produced labels programmatically. Subject-matter experts provide step-by-step guidance for each node's logic. We implement this logic, verify the implementation with them, and run the resulting offline labeler over the historical cases.

The labeler is not itself deployable as a production tool. It is written against dashboard data logged into a data lake of historical snapshots, whereas production tools query live MCPs with different schemas. Historical snapshots are complete, whereas live MCP data may arrive incomplete or with gaps (nulls, missing rows, partial windows), so the tools must be robust to varying patterns of missing data. Production tools also retrieve per-site thresholds from the runtime parameter store and return the observed value, threshold, and explanation alongside the verdict. The pipeline's task is to synthesize implementations that meet this production contract and reproduce the reference verdicts. We expect the pipeline's advantage over hand-written tools to lie less in the initial implementation than in maintenance: the upstream inconsistencies in Section~\ref{sec:lessons} were corrected by rerunning the pipeline, whereas hand-maintained tools would consume engineering time with every environment change, and this cost scales with the size of the tool library.

\section{Further Method Ablations}
\label{sec:appG}

\subsection{Impact of Repair Loop}
\label{sec:fixloop}
The deployed pipeline's repair loop uses single-path reflection, where the model produces a brief diagnosis of the failures and writes a corrected tool after each test round. This is repeated for three rounds. We compare this strategy against two alternatives: no reflection (the model repairs from failures alone without a separate diagnosis step), and multi-hypothesis branching. In multi-hypothesis branching, three independent diagnoses are generated per round. Each is tried as a separate repair branch and we keep the lowest-failure branch.

\begin{figure*}[t]
  \centering
  \includegraphics[width=0.6\textwidth]{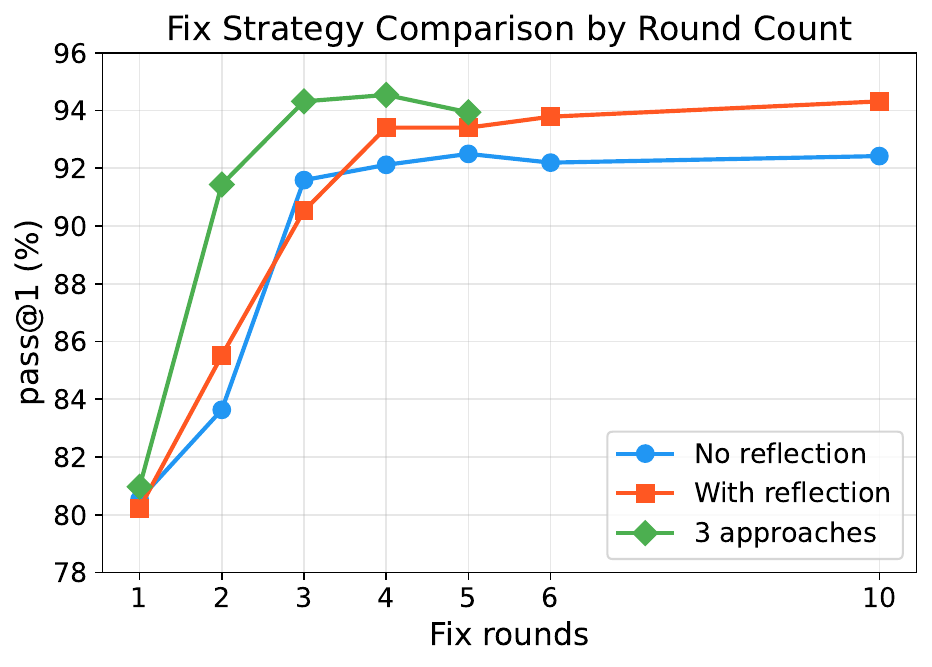}
  \caption{pass@1 across repair rounds for each repair strategy, without the data-collection trace (SOP+R, Opus). No-reflection plateaus around 92.5\%, while reflection peaks at 94.3\% by round 10 but underperforms no-reflection at low round counts. Multi-hypothesis branching reaches 94.3\% by round 3 at roughly $3{\times}$ the per-round cost.}
  \label{fig:rounds}
\end{figure*}

Figure~\ref{fig:rounds} shows that reflection and branching reach similar performance at similar total cost (556K tokens for reflection at 10 rounds, 597K for branching at 4 rounds). Branching reaches the ceiling in fewer rounds because it explores three diagnoses in parallel each round. No-reflection plateaus around 92.5\% because additional rounds repeat the same diagnostic errors.

\subsection{Impact of the Data-Collection Trace}

The two intermediate conditions in Table~\ref{tab:trace} each prepend only a part of the trace to the tool-maker's prompt: the schema metadata observed by the sub-agent (column names, dtypes, row counts), or the sub-agent's verdict graded against the labeled answer.

\begin{table*}[t]
  \centering
  \small
  \begin{tabular}{lcl}
    \toprule
    Trace content & pass@1 & Prompt content \\
    \midrule
    No trace & 91.6 & SOP only \\
    Metadata only & 93.4 & Cols, shapes, rows \\
    Grading only & 93.3 & Determination, label \\
    Full trace & 94.5 & Code, MCP, grading \\
    \bottomrule
  \end{tabular}
  \caption{Trace-content ablation (Opus, three repair rounds). Each component contributes a consistent gain. Confidence intervals are in Table~\ref{tab:a2}.}
  \label{tab:trace}
\end{table*}

Schema metadata adds $+1.8$pp and grading the sub-agent's verdict adds $+1.7$pp, whereas the full trace adds another $+1.2$pp by showing the tool-maker working code. This is a form of execution-grounded feedback that prior work has identified as a significant contributor to repair quality \citep{olausson2024self}.

Two concrete failure modes illustrate the gain. Including the grading corrects semantic biases that are underspecified by the SOP. One tool should return true on zero violations, and models consistently emit a \verb|count > 0| guard. When the sub-agent makes the same error on a sampled case, the trace shows the sub-agent's verdict marked incorrect, and the tool-maker drops the guard in subsequent generations. Including the code can help prevent boundary bugs. Zero-shot tool-maker code often rounds the outcome before comparing to the threshold, whereas the sub-agent's code does not exhibit this failure mode. 

\subsection{Interaction Between the Repair Loop and the Trace}

\begin{table*}[t]
  \centering
  \small
  \begin{tabular}{lccc}
    \toprule
    Strategy (3 rounds) & No trace & Data + grading & $\Delta$ \\
    \midrule
    No reflection & 91.6 & 92.8 & $+1.2$ \\
    With reflection & 90.5 & 94.5 & $+4.0$ \\
    Three hypotheses & 94.3 & 94.4 & $+0.1$ \\
    \bottomrule
  \end{tabular}
  \caption{Interaction between data grounding and repair strategy (Opus, 3 rounds). pass@1 (\%). Confidence intervals and $t$-tests are in Table~\ref{tab:a3}.}
  \label{tab:interact}
\end{table*}

Table~\ref{tab:interact} reports each repair-loop strategy at three repair rounds, the deployed budget, with and without the data-collection trace. The main finding is that the trace yields the largest gain under reflection ($+4.0$pp). At three rounds without the trace, reflection underperforms no-reflection (90.5\% vs.\ 91.6\%), though reflection eventually catches up as shown in Appendix~\ref{sec:fixloop}. The mechanism is that the trace's working code and graded verdict ground the diagnosis step in concrete failure modes, so reflection no longer expends repair rounds on incorrect hypotheses. No-reflection benefits less (91.6\% to 92.8\%, $+1.2$pp) because without an explicit diagnosis step, it does not fully exploit the information in the trace. Branching is essentially unchanged (94.3\% to 94.4\%). With three independent diagnoses, branching explores enough alternatives to converge even without the data-collection trace. With the trace, reflection (94.5\%, 395K tokens) and branching (94.4\%, 496K tokens) reach the same accuracy, but reflection costs 20\% fewer tokens, so the deployed pipeline uses reflection. 

\subsection{Data Efficiency}

\begin{figure*}[t]
  \centering
  \includegraphics[width=0.6\textwidth]{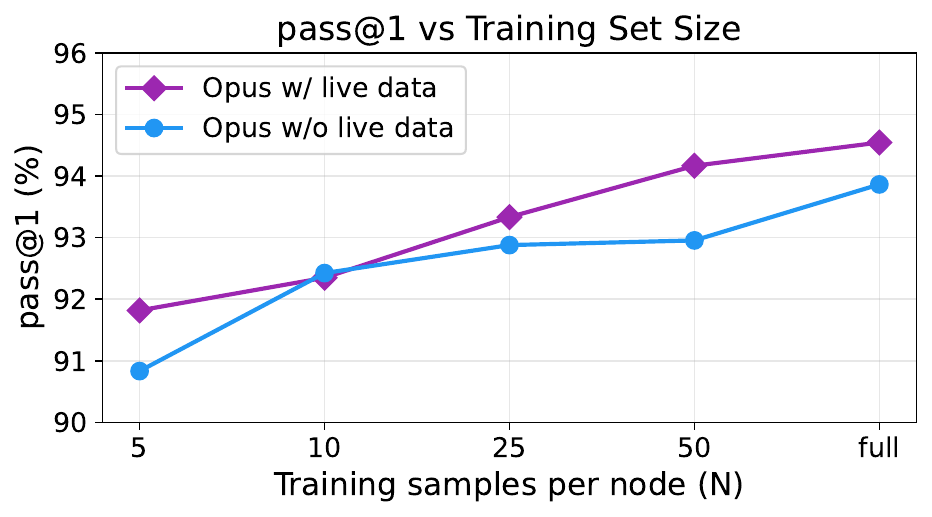}
  \caption{pass@1 vs.\ training-set size, with and without data grounding (Opus).}
  \label{fig:size}
\end{figure*}

Each new SOP node requires labeled training cases to generate its tool. Hence, a practical pipeline must reach high pass@1 from few labels. With grounding, pass@1 reaches 91.8\% at $N{=}5$ training cases per node, within 3pp of the full-set ceiling (Figure~\ref{fig:size}). Removing labels entirely is also feasible. We investigate a label-free tool-maker that uses judge-filtered pseudo-labels and multi-candidate voting instead of the labeled repair-loop test cases. This configuration reaches 91.6\% pass@1, also within 3pp of the labeled ceiling (Appendix~\ref{sec:appC}).

\subsection{Token Scaling}

We perform two follow-ups to the main-text GLM-4.7 ablation (Figure~\ref{fig:cost}). First, we investigate whether GLM-4.7 repair-only catches up to the deployed configuration at higher token budgets. Second, we report the same ablation on a stronger tool-maker, Opus 4.6.

\paragraph{SOP+R does not scale with tokens on GLM-4.7.}
Table~\ref{tab:appG_fixonly} compares the best SOP+R configuration we ran against a higher-budget SOP+R configuration that explores three repair branches per round. Despite spending 55\% more tokens, branching reduces pass@1 from 89.3\% to 87.0\%. The cheapest SOP+D+R configuration (1 sample, 3-round repair) reaches 91.2\% at 596K tokens, which is fewer tokens than the best SOP+R point. The deployed SOP+D+R configuration in the main text reaches 94.5\% at 799K. No SOP+R configuration we ran exceeds 89.3\%.

\begin{table*}[t]
  \centering
  \small
  \begin{tabular}{lcc}
    \toprule
    Configuration & Tokens (K) & pass@1 \\
    \midrule
    SOP+R, 4 rounds                       & 635 & 89.3 \\
    SOP+R, 3 rounds $\times$ 3 branches   & 982 & 87.0 \\
    \midrule
    SOP+D+R, 1 sample, 3 rounds           & 596 & 91.2 \\
    SOP+D+R, 3 samples, 3 rounds          & 799 & 94.5 \\
    \bottomrule
  \end{tabular}
  \caption{GLM-4.7 SOP+R configurations at higher token budgets, paired with two SOP+D+R configurations for reference. +D adds the data-collection trace; +R adds the test-repair loop. Per-seed tokens are means over 30 seeds (10 for the 3-branches row).}
  \label{tab:appG_fixonly}
\end{table*}

\begin{figure*}[t]
  \centering
  \includegraphics[width=0.6\textwidth]{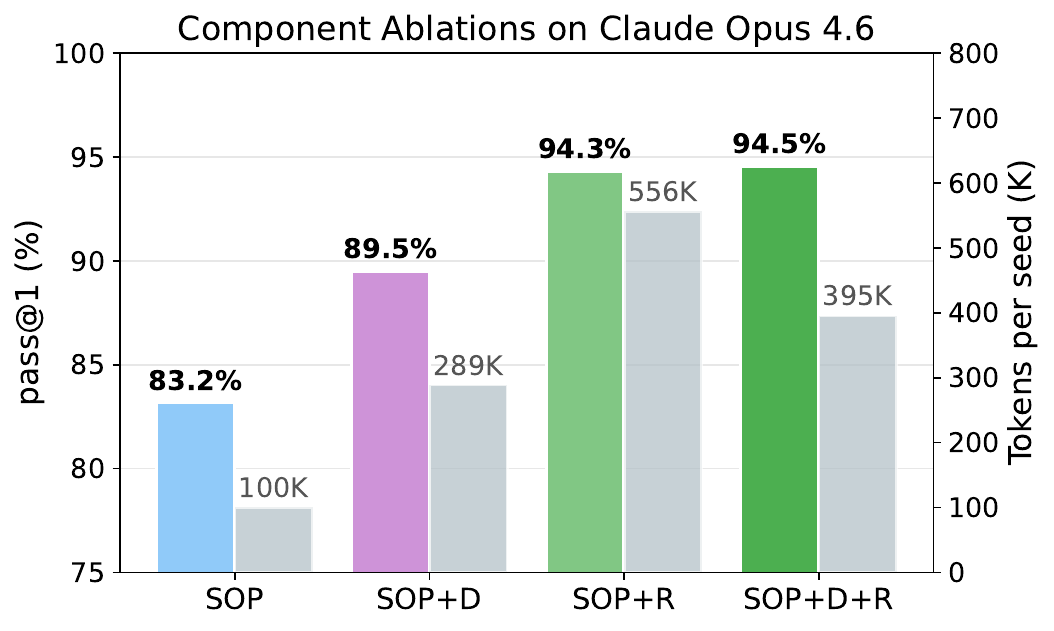}
  \caption{Per-node pass@1 (left bars) and generation cost in tokens per seed (right bars) for each ablation configuration on Claude Opus 4.6. The deployed configuration matches the repair-only ceiling within seed variance at 29\% fewer tokens.}
  \label{fig:appG_opus}
\end{figure*}

\paragraph{Opus 4.6 component ablation.}
Figure~\ref{fig:appG_opus} repeats the Figure~\ref{fig:cost} ablation on Opus 4.6. The repair loop alone reaches 94.3\% pass@1 at 556K tokens with 10 rounds. The deployed configuration reaches comparable accuracy (94.5\%) at 395K tokens, a 29\% token reduction. The trace helps Opus reach the ceiling at lower cost, but the ceiling is unchanged.

\section{Label-Free Generation}
\label{sec:appC}

We investigate how to replace the labeled repair loop with a pipeline that combines voting and LLM judges. For each tool, $K$ independent sub-agents each investigate one sampled unlabeled case. An LLM judge filters out traces that are not consistent with the SOP. From the surviving traces, we generate $N$ independent candidates, and execute each candidate against the $K$ sampled cases. We then assign a pseudo-label for each case by majority vote across the surviving traces and the $N$ candidate outputs, and iterate a repair loop for $R$ rounds. Each round, every candidate is executed on the $K$ investigated cases. Candidates that agree with all current pseudo-labels and pass a code-review judge (an LLM that inspects the candidate tool's code for consistency with the SOP) are carried forward unchanged. The remaining candidates see their disagreement cases as failure feedback, write a short diagnosis, and regenerate. After all $N$ candidates have either regenerated or been carried forward, the pseudo-labels for each case are recomputed by majority vote over the updated candidate outputs and the judge-passing sub-agent determinations. The final candidate is selected by counting how many of the $K$ investigated cases each candidate matches the pseudo-label on. Ties are broken first by the code-review judge, then by code length.

The headline configuration is GLM-4.7 as the tool-maker, Claude Sonnet 4.5 as the trace judge, and Claude Opus 4.7 as the reflector agent, with $K{=}3$, $N{=}10$, and $R{=}5$. The reflector here is Opus 4.7 due to model availability at the time of this experiment. We do not attribute the result to the newer model. This reaches 91.6\% pass@1 (cf.\ the 94.5\% labeled ceiling). The most significant component is voting across $N$ candidates: $N{=}1$ loses 8.6pp. Other ablations (removing the repair loop, reducing to $K{=}1$, skipping the code-review judge) each cost 2.1--3.4pp. Once we clarify the specification as in Section~\ref{sec:spec}, Opus 4.7 without labels reaches 99.1\% pass@1.

\section{The Repair Loop Unlearns Correct Code}
\label{sec:appD}

The fourth residual node in Section~\ref{sec:spec} is a precondition for a downstaffing action: downstaffing only proceeds if the collector belt is sufficiently empty. The SOP requires, for each quadrant, a 5-minute average leading up to that quadrant's most recent timestamp, with no rows older than 10 minutes included. The round-0 candidate below is mostly correct, but uses $>$ instead of the SOP's $\ge$ at the start boundary, which produces 6 test failures in round 1. The round-2 candidate replaces the 5-minute averaging logic with a simple average over the full 10-minute window, which produces fewer test failures, but no longer matches the SOP.

\vspace{5mm}

\begin{minipage}{0.9\linewidth}
\begin{lstlisting}[language=Python,basicstyle=\scriptsize\ttfamily]
# Round 0: per-quadrant 5-minute window
# (uses > instead of the SOP's implied >=)
fresh = df[
    (df['five_minute_interval'] >= start_time) &
    (df['five_minute_interval'] <= current_time)
]
for q, col in quadrants.items():
    q_data = fresh[fresh[col].notna()]
    analysis_time = q_data['five_minute_interval'].max()
    five_min_start = max(analysis_time - timedelta(minutes=5), start_time)
    subset = q_data[
        (q_data['five_minute_interval'] >  five_min_start) &
        (q_data['five_minute_interval'] <= analysis_time)
    ]
    avgs[q] = subset[col].mean()

# Round 2: 5-minute window collapsed to the
# full 10-minute window per quadrant
fresh = df[
    (df['five_minute_interval'] >= start_time) &
    (df['five_minute_interval'] <= current_time)
]
for q, col in quadrants.items():
    avgs[q] = fresh[col].dropna().mean()

# Both rounds: thresholding and outcome (unchanged)
north_avg = (avgs['NW'] + avgs['NE']) / 2
south_avg = (avgs['SW'] + avgs['SE']) / 2
holds = (north_avg <= north_threshold and
         south_avg <= south_threshold)
\end{lstlisting}
\end{minipage}

\FloatBarrier
\clearpage

\section{Per-Component Output-Token Breakdown}
\label{sec:appE}

Section~\ref{sec:online} reports a 58\% reduction in per-alarm output tokens with tools. Table~\ref{tab:e2} gives a breakdown by agent role (main vs.\ sub-agent) and by content type (reasoning, prose text, code blocks). The main contributor to the reduction is sub-agent code generation ($-80\%$), which is expected when per-invocation code generation is replaced with tool calls. The sub-agent also needs to reason less, as calling the tool is simpler than writing the code. The main agent's tokens also fall ($-16\%$), as reading a structured verdict is cheaper than parsing the sub-agent's free-text answer. The 20\% of sub-agent code tokens that remain consist mostly of the tool calls themselves, which appear as code blocks under the CodeAct-style execution model, and the agent's programmatic to-do list.

\begin{table}[h]
  \centering
  \small
  \begin{tabular}{lrrr}
    \toprule
    Component & With tools & No tools & $\Delta$ \\
    \midrule
    Main: thinking & 1{,}118 & 1{,}349 & $-17\%$ \\
    Main: text & 391 & 436 & $-10\%$ \\
    Main: code & 202 & 259 & $-22\%$ \\
    \textbf{Main total} & \textbf{1{,}711} & \textbf{2{,}044} & $-\mathbf{16\%}$ \\
    Sub: thinking & 2{,}013 & 4{,}144 & $-51\%$ \\
    Sub: text & 1{,}035 & 1{,}358 & $-24\%$ \\
    Sub: code & 1{,}466 & 7{,}452 & $-80\%$ \\
    \textbf{Sub total} & \textbf{4{,}514} & \textbf{12{,}954} & $-\mathbf{65\%}$ \\
    \textbf{Total} & \textbf{6{,}225} & \textbf{14{,}998} & $-\mathbf{58\%}$ \\
    \bottomrule
  \end{tabular}
  \caption{Per-trace output-token breakdown by agent role and content type, per-alarm production means (Qwen3 32B).}
  \label{tab:e2}
\end{table}

\clearpage
\FloatBarrier

\newpage
\section{Confidence Intervals and Significance Tests}
\label{sec:appA}

Point estimates reported elsewhere in the paper are repeated here with 95\% confidence intervals (mean $\pm 1.96 \times$ SE over per-seed pass@1 scores) and corresponding significance tests.

\begin{table*}[h]
  \centering
  \small
  \begin{tabular}{lcccc}
    \toprule
     & SOP & SOP+D & SOP+R & SOP+D+R \\
    \midrule
    GLM-4.7 & 75.7 [74.5, 76.8] & 84.9 [83.5, 86.2] & 89.3 [87.8, 90.8] & 94.5 [93.8, 95.1] \\
    GLM-5 & 76.8 [75.4, 78.3] & 83.4 [82.0, 84.8] & 88.0 [87.0, 89.0] & 93.6 [92.8, 94.3] \\
    Qwen3 235B & 74.2 [72.3, 76.2] & 71.6 [70.2, 72.9] & 79.9 [78.9, 80.9] & 81.1 [79.9, 82.2] \\
    GLM-4.7 Flash & 56.4 [54.4, 58.5] & 60.4 [59.2, 61.6] & 58.7 [57.7, 59.8] & 65.8 [64.3, 67.3] \\
    \bottomrule
  \end{tabular}
  \caption{Component ablations of the deployed pipeline with 95\% CIs (pass@1, \%), corresponding to the point estimates in Table 1. +D adds the data-collection trace (Section~\ref{sec:gen}); +R adds the test-repair loop. SOP+D+R is the deployed configuration. Differences between SOP+D+R and the SOP+D and SOP+R ablations are statistically significant on GLM-4.7, GLM-5, and GLM-4.7 Flash (significance tests in Table~\ref{tab:a5}).}
  \label{tab:paradigms-ci}
\end{table*}

\begin{table*}[h]
  \centering
  \small
  \begin{tabular}{lc}
    \toprule
    Trace content & pass@1 \\
    \midrule
    No trace & 91.6 [90.7, 92.5] \\
    Metadata only & 93.4 [92.9, 93.9] \\
    Grading only & 93.3 [92.4, 94.3] \\
    Full trace & 94.5 [93.8, 95.3] \\
    \bottomrule
  \end{tabular}
  \caption{Trace-content ablation with CIs (Table~\ref{tab:trace}). Opus, three repair rounds.}
  \label{tab:a2}
\end{table*}

\begin{table*}[h]
  \centering
  \small
  \begin{tabular}{lcccrr}
    \toprule
    Strategy & No trace & Data + grading & $\Delta$ & $t$ & $p$ \\
    \midrule
    No reflection & 91.6 [90.7, 92.5] & 92.8 [92.2, 93.4] & $+1.21$ & 2.50 & 0.018 \\
    With reflection & 90.5 [89.5, 91.5] & 94.5 [93.8, 95.3] & $+4.01$ & 6.25 & ${<}10^{-6}$ \\
    Three hypotheses & 94.3 [93.3, 95.3] & 94.4 [93.6, 95.2] & $+0.08$ & 0.11 & 0.91 \\
    \bottomrule
  \end{tabular}
  \caption{Data grounding and repair strategy with CIs and paired $t$-tests (Table~\ref{tab:interact}). Opus, three repair rounds, 30 seeds.}
  \label{tab:a3}
\end{table*}

\paragraph{Per-seed pass@1 with confidence intervals.} Tables~\ref{tab:paradigms-ci}, \ref{tab:a2}, and~\ref{tab:a3} give the build-time pipeline results from Section~\ref{sec:offline} and Appendix~\ref{sec:appG} with 95\% CIs and paired $t$-tests over 30 seeds (10 for the SOP column).

\clearpage

\begin{table}[H]
  \centering
  \small
  \begin{tabular}{lrrrcc}
    \toprule
    Metric & With & Without & $\Delta$ & 95\% CI & $p$ \\
    \midrule
    Obs acc. & 99.62 & 99.71 & $-0.09$ & [$-0.38$, $+0.18$] & 0.57 \\
    RC acc. & 100.00 & 99.33 & $+0.67$ & [$+0.33$, $+1.04$] & ${<}10^{-3}$ \\
    Act acc. & 98.38 & 97.44 & $+0.93$ & [$+0.31$, $+1.58$] & 0.0024 \\
    Overall acc. & 98.18 & 97.18 & $+1.00$ & [$+0.31$, $+1.69$] & 0.0052 \\
    \bottomrule
  \end{tabular}
  \caption{Qwen3 32B, with vs.\ without tools (\%). 95\% CIs and $p$-values are bootstrap-by-task ($n_{\text{boot}}=5{,}000$).}
  \label{tab:a4qwen}
\end{table}

\begin{table}[H]
  \centering
  \small
  \begin{tabular}{lrrrcc}
    \toprule
    Metric & With & Without & $\Delta$ & 95\% CI & $p$ \\
    \midrule
    Obs acc. & 99.87 & 99.18 & $+0.69$ & [$+0.38$, $+1.00$] & ${<}10^{-3}$ \\
    RC acc. & 99.71 & 99.49 & $+0.22$ & [$-0.04$, $+0.51$] & 0.12 \\
    Act acc. & 99.36 & 99.29 & $+0.07$ & [$-0.22$, $+0.38$] & 0.70 \\
    Overall acc. & 99.24 & 98.29 & $+0.96$ & [$+0.51$, $+1.42$] & ${<}10^{-3}$ \\
    \bottomrule
  \end{tabular}
  \caption{GLM-4.5-Air, with vs.\ without tools (\%). 95\% CIs and $p$-values are bootstrap-by-task ($n_{\text{boot}}=5{,}000$).}
  \label{tab:a4glm}
\end{table}

\begin{table}[H]
  \centering
  \small
  \begin{tabular}{lrrrcc}
    \toprule
    Metric & Main-agent direct & Without & $\Delta$ & 95\% CI & $p$ \\
    \midrule
    Obs acc. & 99.91 & 99.18 & $+0.73$ & [$+0.44$, $+1.04$] & ${<}10^{-3}$ \\
    RC acc. & 99.98 & 99.49 & $+0.49$ & [$+0.24$, $+0.76$] & ${<}10^{-3}$ \\
    Act acc. & 99.40 & 99.29 & $+0.11$ & [$-0.22$, $+0.44$] & 0.57 \\
    Overall acc. & 99.31 & 98.29 & $+1.02$ & [$+0.56$, $+1.49$] & ${<}10^{-3}$ \\
    \bottomrule
  \end{tabular}
  \caption{GLM-4.5-Air with main-agent direct tool calls vs.\ without tools (\%). 95\% CIs and $p$-values are bootstrap-by-task ($n_{\text{boot}}=5{,}000$).}
  \label{tab:a4glmdirect}
\end{table}

\paragraph{Significance of end-to-end correctness (Table~\ref{tab:arch}).} For each model, we run three seeds with tools and three seeds without tools, and we evaluate each seed on the same 1{,}500 tasks. To compute confidence intervals and $p$-values, we aggregate at the task level because we cannot treat the 4{,}500 (task, seed) pairs as independent. We average the three seed-level results per task, and we resample tasks with replacement. The reported 95\% CIs are the 2.5th and 97.5th percentiles of the resampled accuracy difference (with tools minus without tools) over $n_{\text{boot}}=5{,}000$ bootstrap iterations. The reported $p$-value is the fraction of bootstrap iterations in which the resampled accuracy difference was zero or had the opposite sign to the observed delta, multiplied by two for the two-sided test.

Improvements in overall accuracy are significant for both agent models in the sub-agent architecture (Tables~\ref{tab:a4qwen}~and~\ref{tab:a4glm}) and persist when the main agent calls tools directly (Table~\ref{tab:a4glmdirect}). Qwen3 32B's gains come primarily from improved root-cause and action accuracy, whereas GLM-4.5-Air analyzes observations more accurately.  

\FloatBarrier
\clearpage

\begin{table*}[t]
  \centering
  \small
  \begin{tabular}{lrrr}
    \toprule
    Comparison & $\Delta$ & $t$ & $p$ \\
    \midrule
    GLM-4.7: SOP+R vs.\ SOP+D & $+4.46$ & 4.60 & $7.7{\times}10^{-5}$ \\
    GLM-4.7: SOP+D+R vs.\ SOP+R & $+5.15$ & 6.32 & $6.6{\times}10^{-7}$ \\
    GLM-4.7: SOP+D+R vs.\ SOP+D & $+9.62$ & 13.58 & $4.2{\times}10^{-14}$ \\
    GLM-5: SOP+R vs.\ SOP+D & $+4.57$ & 4.78 & $4.6{\times}10^{-5}$ \\
    GLM-5: SOP+D+R vs.\ SOP+R & $+5.55$ & 7.52 & $2.7{\times}10^{-8}$ \\
    GLM-5: SOP+D+R vs.\ SOP+D & $+10.12$ & 13.21 & $8.5{\times}10^{-14}$ \\
    Qwen3 235B: SOP+R vs.\ SOP+D & $+8.34$ & 10.38 & $2.8{\times}10^{-11}$ \\
    Qwen3 235B: SOP+D+R vs.\ SOP+R & $+1.14$ & 1.72 & $0.096$ \\
    Qwen3 235B: SOP+D+R vs.\ SOP+D & $+9.48$ & 11.42 & $3.0{\times}10^{-12}$ \\
    GLM-4.7 Flash: SOP+R vs.\ SOP+D & $-1.66$ & $-2.12$ & $0.042$ \\
    GLM-4.7 Flash: SOP+D+R vs.\ SOP+R & $+7.05$ & 6.78 & $1.9{\times}10^{-7}$ \\
    GLM-4.7 Flash: SOP+D+R vs.\ SOP+D & $+5.39$ & 5.38 & $8.8{\times}10^{-6}$ \\
    \bottomrule
  \end{tabular}
  \caption{Significance of Table~\ref{tab:paradigms-ci} ablation differences. +D adds the data-collection trace; +R adds the test-repair loop. Paired $t$-tests over 30 seeds. $\Delta$ in pp.}
  \label{tab:a5}
\end{table*}

\paragraph{Significance of repair loop and trace contributions.} For the two best models, GLM-4.7 and GLM-5, all three pairwise comparisons are significant ($p < 10^{-4}$). The deployed configuration (SOP+D+R) is significantly better than either single-component ablation, and both components contribute. On Qwen3 235B, SOP+D+R is not significantly better than SOP+R alone. Manual inspection of regressions shows that Qwen3 235B incorrectly infers schema details from the data-collection trace (wrong tool selection, wrong column types, missing columns), producing tools that crash or return "No data" on inputs the SOP-only configuration handles correctly. On GLM-4.7 Flash, the smaller tool-maker fails to productively use the repair loop without the trace to ground the repairs. The performance of SOP+D+R exceeds that of SOP+R and SOP+D, and neither component alone matches the gain of the combined configuration.

\begin{table*}[t]
  \centering
  \small
  \begin{tabular}{lrrrrr}
    \toprule
    Comparison & $\Delta$ & $t$ & $p$ & $n_x$ & $n_y$ \\
    \midrule
    Aug.\ data vs.\ baseline      & $+2.05$ & 3.25  & $4.1{\times}10^{-3}$  & 10 & 30 \\
    Clarified SOP vs.\ aug.\ data & $+3.33$ & 6.49  & $9.3{\times}10^{-5}$  & 30 & 10 \\
    Clarified SOP vs.\ baseline   & $+5.38$ & 14.22 & $3.1{\times}10^{-15}$ & 30 & 30 \\
    \bottomrule
  \end{tabular}
  \caption{Significance of the specification ceiling interventions (Table~\ref{tab:ceiling}). Two-sample Welch's $t$-tests over per-seed pass@1. $\Delta$ in pp. $n_x, n_y$ are the per-condition seed counts.}
  \label{tab:a6}
\end{table*}

\paragraph{Significance of specification interventions.} Both interventions, clarifying the SOP and augmenting the training data, are individually significant (Table~\ref{tab:a6}). 

\end{document}